\documentclass[letterpaper, 10 pt, conference]{ieeeconf}  % Comment this line out if you need a4paper

\IEEEoverridecommandlockouts              % This command is only needed if 
\makeatletter
\let\NAT@parse\undefined
\makeatother
\usepackage{graphicx} % for pdf, bitmapped graphics files
\usepackage{amsmath} % assumes amsmath package installed
\usepackage{amssymb}  % assumes amsmath package installed
\usepackage{lipsum}
\usepackage{caption}
\usepackage{subcaption}
\usepackage{cite}

\usepackage{bm}
\usepackage{multirow}
\newcommand*{\B}[1]{\ifmmode\bm{#1}\else\textbf{#1}\fi}

\makeatletter
\newcommand{\subalign}[1]{%
  \vcenter{%
    \Let@ \restore@math@cr \default@tag
    \baselineskip\fontdimen10 \scriptfont\tw@
    \advance\baselineskip\fontdimen12 \scriptfont\tw@
    \lineskip\thr@@\fontdimen8 \scriptfont\thr@@
    \lineskiplimit\lineskip
    \ialign{\hfil$\m@th\scriptstyle##$&$\m@th\scriptstyle{}##$\hfil\crcr
      #1\crcr
    }%
  }%
}
\makeatother

\usepackage{amsmath}
\usepackage{amssymb}
\usepackage{booktabs}
\usepackage{multirow}
\usepackage{url}
\usepackage{bm}
\usepackage{makecell}
\usepackage {colortbl,array,xcolor}
\renewcommand{\arraystretch}{1.2}

\def\red#1{\textcolor{red}{\textbf{#1}}}
\def\blue#1{\textcolor{blue}{\textbf{#1}}}

\definecolor{rasred}{rgb}{0.596,0.003,0.180}
\definecolor{pltgreen}{rgb}{0.0, 0.502, 0.0}
\def\pltgreen#1{\textcolor{pltgreen}{\textbf{#1}}}
\usepackage[pagebackref=true,breaklinks=true,letterpaper=true,colorlinks,bookmarks=false,citecolor=rasred]{hyperref}

\title{\LARGE \bf
Robot Motion Planning using One-Step Diffusion\\ with Noise-Optimized Approximate Motions
}
\author{Tomoharu Aizu$^{1}$, Takeru Oba$^{1}$, Yuki Kondo$^{1}$, and Norimichi Ukita$^{1*}$
\thanks{$^*$ {\tt\small ukita@toyota-ti.ac.jp}}
\thanks{$^{1}$Toyota Technological Institute, Nagoya, Aichi, Japan.}%
}

\begin{document}

\maketitle
\global\csname @topnum\endcsname 0
\global\csname @botnum\endcsname 0
\thispagestyle{empty}
\pagestyle{empty}

%%%%%%%%%%%%%%%%%%%%%%%%%%%%%%%%%%%%%%%%%%%%%%%%%%%%%%%%%%%%%%%%%%%%%%%%%%%%%%%%
\begin{abstract}
This paper proposes an image-based robot motion planning method using a one-step diffusion model.
While the diffusion model allows for high-quality motion generation, its computational cost is too expensive to control a robot in real time.
To achieve high quality and efficiency simultaneously, our one-step diffusion model takes an approximately generated motion, which is predicted directly from input images.
This approximate motion is optimized by additive noise provided by our novel noise optimizer.
Unlike general isotropic noise, our noise optimizer adjusts noise anisotropically depending on the uncertainty of each motion element.
Our experimental results demonstrate that our method outperforms state-of-the-art methods while maintaining its efficiency by one-step diffusion.

\end{abstract}

%%%%%%%%%%%%%%%%%%%%%%%%%%%%%%%%%%%%%%%%%%%%%%%%%%%%%%%%%%%%%%%%%%%%%%

\section{Introduction}
\label{section: introduction}

For robot motion planning, we have to compare the current state of a robot with its surrounding environment.
Among various sensors for observing the environment, including the robot, optical sensors such as RGB and RGB-Depth cameras are widely used because of their wide availability, wide observation ranges, and so on.
We call robot motion planning using camera images {\em image-based robot motion planning}~\cite{read, Q-attention, Diffusion_Policy}.
For robot motion planning, including image-based robot motion planning, reinforcement learning~\cite{Q-attention, seo2023masked} and imitation learning~\cite{Diffusion_Policy, IBC} are two major approaches.

While reinforcement learning is potentially powerful, its difficulty in designing reward functions makes it impractical to apply reinforcement learning to various real-world scenarios.
In imitation learning, on the other hand, demonstration motions that achieve a task are given as training data instead of the reward functions in reinforcement learning.
Such demonstration collection becomes easier than before by recent advances in robotics hardware~\cite{Mobile_aloha} and simulators~\cite{Scalingup}.

\begin{figure}[t]
\begin{center}
    \includegraphics[width=\columnwidth]{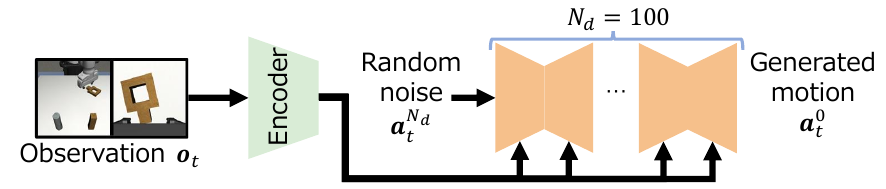}\\
    \vspace*{0mm}
    (a) Diffusion policy~\cite{Diffusion_Policy}\\
    \vspace*{2mm}
    \includegraphics[width=\columnwidth]{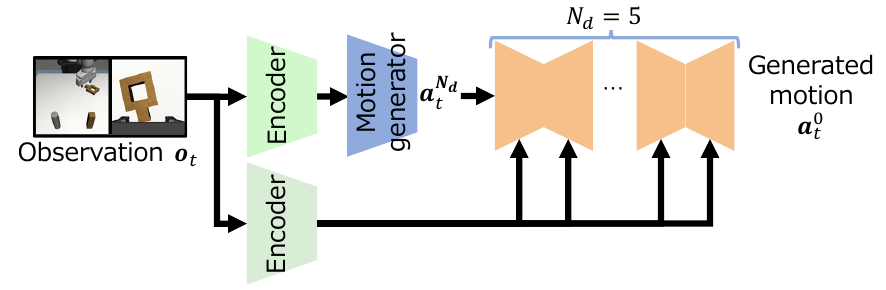}\\
    \vspace*{-2mm}
    (b) BRIDGER~\cite{bridger_chen2024behavioral}\\
    \vspace*{2mm}
    \includegraphics[width=\columnwidth]{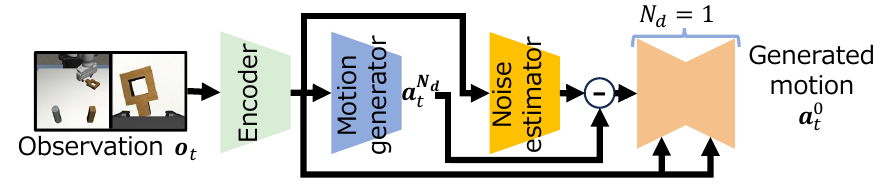}\\
    (c) Our proposed method\\
    \caption{Previous methods~\cite{Diffusion_Policy,bridger_chen2024behavioral} vs. our proposed method.}
    \label{fig: intro}
\end{center}
\vspace*{-2mm}
\end{figure}

While this advantage of imitation learning motivates us to propose various imitation learning-based robot motion planning methods~\cite{robomimic2021,shafiullah2022behavior,IBC}, their performance is still limited due to the insufficient learning capability of their backbone machine learning models.
On the other hand, diffusion models~\cite{ddpm} allow us to improve various generative tasks, such as image generation~\cite{text-to-image_ZhaoRLLZL23,Wang2024TokenCompose} and human motion generation~\cite{dance_ZhangTZLHX024}, including robot motion planning~\cite{Diffusion_Policy,read}.
In diffusion models, an input sample (e.g., motion data) is iteratively denoised (i.e., refined) towards the final output.
In early motion planning methods using diffusion models~\cite{Diffusion_Policy}, a huge number of steps (e.g., between 100 and 1,000) are required to generate accurate motion trajectories (i.e., motions that can successfully achieve a given task), as shown in Fig.~\ref{fig: intro} (a).
However, such a huge number of steps is computationally expensive.
On the other hand, in BRIDGER~\cite{bridger_chen2024behavioral} (Fig.~\ref{fig: intro} (b)), which is the prior work closest to our method, the number of steps (denoted by $N_{d}$) is reduced by feeding an approximately accurate motion as an initial motion into the diffusion model.

This paper proposes further reducing the iterative diffusion steps to one step, as shown in Fig.~\ref{fig: intro} (c).
Although inaccurate motions are generated if $N_{d}$ is simply reduced to one, this paper improves an approximately generated initial motion by optimizing diffusion noise given to this initial motion.
While this noise is not optimized
%(i.e., not explicitly given) 
in BRIDGER and other similar methods, the noise level is essential to make the diffusion model work accurately~\cite{Importance_noise_schedule, SDEdit}.
Our method optimizes the noise level depending on the observation by the noise estimator, as shown in Fig.~\ref{fig: intro} (c).

Our contributions are summarized as follows:
\begin{itemize}
\item One-step diffusion is achieved for robot motion generation by improving an initial motion given to the diffusion model so that this initial motion gets close to accurate task-achievable motions.
\item The initial motion is sampled from the noise distribution anisotropically optimized based on the observation.
\end{itemize}

%%%%%%%%%%%%%%%%%%%%%%%%%%%%%%%%%%%%%%%%%%%%%%%%%%%%%%%%%%%%%%%%%%%%%%

\section{Related Work}
\label{section: related}

\subsection{Diffusion Models}

Diffusion models~\cite{ddpm} are used for various generative tasks such as image generation~\cite{text-to-image_ZhaoRLLZL23,Wang2024TokenCompose}, human motion generation~\cite{dance_ZhangTZLHX024}, and robot motion generation~\cite{Diffusion_Policy,read}.
For image-based robot motion planning, we have to control a diffusion model so that its generated motion can achieve a given task in the environment observed in an input image.
Such diffusion control approaches are categorized into early-step skip~\cite{SDEdit,BSRD} and conditioning~\cite{text-to-image_ZhaoRLLZL23,Wang2024TokenCompose}.

In the {\bf early-step skip diffusion} approach, denoising begins not from random noise but from a sample approximating the final denoised output.
For example, for image style transfer~\cite{SDEdit}, early diffusion steps are skipped by feeding a source-style image into an intermediate diffusion step
of the diffusion model.

While an approximate sample whose domain must be the same as the domain of the final output is required for the early-step skip approach, data of any domain can be used in the {\bf conditioning} approach.
For example, in the text-to-image generation task~\cite{text-to-image_ZhaoRLLZL23}, a text prompt is given as the condition to generate an image representing the text's context.

\subsection{Diffusion Models for Motion Generation}

The aforementioned early-step skip and conditioning approaches are also used for robot motion planning.
With the early-step skip approach, READ~\cite{read} and BRIDGER~\cite{bridger_chen2024behavioral} use motions that are retrieved and approximately generated based on the current states of a robot and its surrounding environment, respectively.
With the conditioning approach~\cite{human_behaviour_PearceRKBSGMTMH23,Diffusion_Policy,crossway_diffusion,ma2024hierarchical,beso_ReussLJL23,diffusion_copolicy_NgLK24,bridger_chen2024behavioral}, the current states of a robot and its surrounding environment are observed by a camera, and the observed image is provided as a condition for the diffusion model to generate a future motion that achieves a given task.
Our method integrates both of these two approaches for further improvement.

\subsection{Diffusion Step Reduction}
\label{subsection: step_reduction}

Since the iterative steps of the diffusion model lead to high computational cost, various approaches are proposed to reduce the cost. % can be removed
In consistency model~\cite{consistencymodels}, a pre-trained diffusion model is distilled to reduce $N_{d}$.
While Consistency Policy~\cite{prasad2024consistency} applies this consistency model to Diffusion Policy~\cite{Diffusion_Policy} for robot motion planning, its performance is limited in one-step diffusion.
In One-Step Diffusion Policy (OneDP)~\cite{wang2024onestepdiffusionpolicyfast}, additional constraints are used for the distillation process to improve the performance.
While these methods using distillation~\cite{prasad2024consistency,wang2024onestepdiffusionpolicyfast} require a two-step training strategy (i.e., (1) pre-training of a teacher model and (2) distillation from the teacher model), Shortcut Model~\cite{frans2024onestepdiffusionshortcutmodels} enables one-step diffusion with a one-step training process.
However, all the methods introduced in Sec.~\ref{subsection: step_reduction} are still inferior to Diffusion Policy~\cite{Diffusion_Policy} in terms of motion accuracy.

%%%%%%%%%%%%%%%%%%%%%%%%%%%%%%%%%%%%%%%%%%%%%%%%%%%%%%%%%%%%%%%%%%%%%%

\section{Problem Definition and Notation}
\label{section: notation}

\begin{figure}[t]
\begin{center}
    \includegraphics[width=0.9\columnwidth]{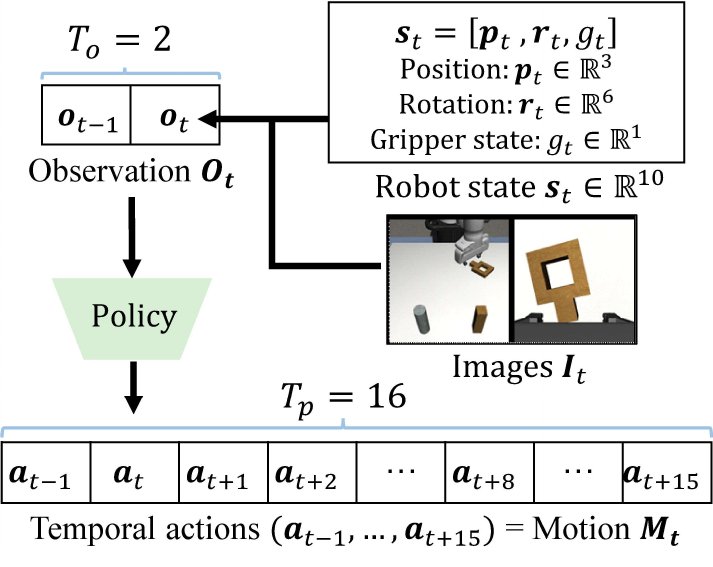}\\
    \caption{Input and output data in our method.}
    \label{fig: variable}
\end{center}
\vspace*{-2mm}
\end{figure}

Figure~\ref{fig: variable} illustrates the input and output data of our method.
The observation at time step $t$ is denoted as $\bm{o}_t$, which consists of the robot state $\bm{s}_t$ and RGB images $\bm{I}_t \in \mathbb{R}^{H \times W \times C \times N_{i}}$ where $N_{i}$ denotes the number of input images.
Next, the action at $t$ is denoted as $\bm{a}_t$. In our experiments, both $\bm{s}_t$ and $\bm{a}_t$ are $\mathbb{R}^{10}$, consisting of the 3D position $\bm{p}_t \in \mathbb{R}^3$, the orientation with continuous representation~\cite{zhou2019continuity} $\bm{r}_t \in \mathbb{R}^6$, and the gripper state $g_t \in \mathbb{R}^1$ of a robot hand. 

Predicting $\bm{a}_t$ from $\bm{o}_t$ is common in imitation learning. Its recent studies~\cite{Diffusion_Policy, bridger_chen2024behavioral} reveal that incorporating multi-frame observations and predictions improves prediction accuracy. Following them~\cite{Diffusion_Policy, bridger_chen2024behavioral}, our method employs multi-frame observations and actions. We define multi-frame actions as motion in this paper. As shown in Fig.~\ref{fig: variable}, the observation and prediction lengths are denoted as $T_o$ and $T_p$, and the multi-frame observations and motion as $\bm{O}_t$ and $\bm{M}_t$, respectively.

In the diffusion model, the motion $\bm{M}_t$ at diffusion step $k$ is denoted as $\bm{M}_t^k$, where $k$ is an integer ranging from $0$ to $K$.
$\bm{M}_t^K$ represents the most noisy motion, while $\bm{M}_t^0$ corresponds to the cleanest motion.

\section{NO-Diffusion: One-step Diffusion with Noise-Optimized Approximate Motion}
\label{section: method}

\begin{figure}[t]
\begin{center}
    \includegraphics[width=1.0\columnwidth]{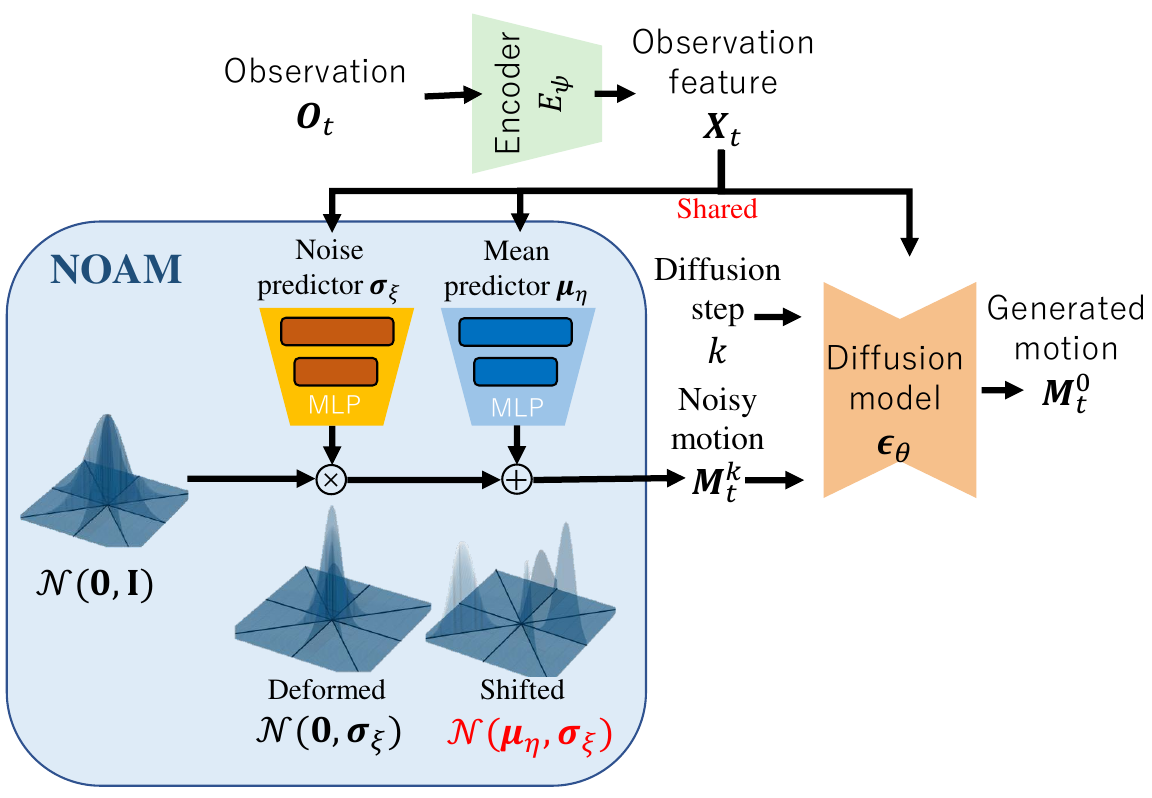}\\
    \caption{Overview of our method, NO-Diffusion, and the detail of our proposed Noise-optimized Approximate Motion Estimator, NOAM.}
    \label{fig: method}
\end{center}
\vspace*{-2mm}
\end{figure}

Sections~\ref{subsection: noise-optimized motion estimator} and~\ref{subsection: overview} describe the overview of our method and the details of our novel noise optimizer, respectively.
The training scheme is explained in Sec.~\ref{subsection: training}.

\subsection{Overview}
\label{subsection: overview}

The overview of our noise-optimized one-step diffusion, NO-Diffusion, is illustrated in Fig.~\ref{fig: method}.
At $t$, the observation $\bm{O}_t$ is fed into the observation encoder $E_{\psi}$, producing the observation feature $\bm{X}_t$. 
$\bm{X}_t$, a noisy motion $\bm{M}_t^k$, and the diffusion step $k$ are fed into the diffusion model $\epsilon_{\theta}$.
The reverse diffusion process starts with $\bm{M}_t^K$ to progressively denoise it to $\bm{M}_t^0$.

In standard diffusion methods~\cite{Diffusion_Policy, crossway_diffusion}, $\bm{M}_t^K$ is sampled from a standard Gaussian distribution $\mathcal{N}(0,{\rm \bm{I}})$. However, a huge number of denoising steps are required if $\bm{M}_t^K$ is far from $\bm{M}_t^0$ (e.g., $K=100$ in~\cite{Diffusion_Policy}).
To accelerate this denoising process by reducing $K$, we propose a Noise-optimized Approximate Motion estimator (NOAM) depicted on the left side of Fig.~\ref{fig: method}.
NOAM brings $\bm{M}_t^K$ closer to $\bm{M}_t^0$ by estimating the appropriate mean $\bm{\mu}$ and variance $\bm{\sigma}$ of the noise distribution $\mathcal{N}(\bm{\mu},\bm{\sigma})$ from which $\bm{M}_t^K$ is sampled.
Its detail is explained in Section~\ref{subsection: noise-optimized motion estimator}.
$\epsilon_{\theta}$ denoises $\bm{M}_t^K$. Unlike conventional diffusion models, NOAM allows $\epsilon_{\theta}$ to predict $\bm{M}_t^0$ by one step ($K=1$).

By executing the one-step diffusion process mentioned above at each time step $t$, the overall motion is determined through closed-loop predictions.

\subsection{Noise-optimized Approximate Motion Estimator (NOAM)}
\label{subsection: noise-optimized motion estimator}

NOAM accelerates the denoising process by shifting the distribution of $\bm{M}_t^K$ toward that of $\bm{M}_t^0$.

This distribution shift is estimated in a data-driven manner.
The key requirements to this estimation for real-time processing are (a) estimating a distribution that is easy to sample from and (b) employing a lightweight network. 
%%%

\vskip\baselineskip
\noindent
\textbf{(a) Observation-conditioned anisotropic Gaussian:}

NOAM adopts the \textbf{anisotropic Gaussian distribution} as $\mathcal{N}(\bm{\mu},\bm{\sigma})$ for sampling $\bm{M}_t^K$ because it enables motion-elementwise noise optimization.
Unlike common generation tasks (e.g., image generation), a robot motion consists of multiple modalities, such as positions, angles, and gripper states.
Their ranges and distributions differ from each other.
The anisotropic Gaussian distribution is well-suited for estimating such multi-modalities.

Moreover, NOAM optimizes $\mathcal{N}(\bm{\mu},\bm{\sigma})$ depending on observation $\bm{O}_t$. 
This observation-dependent sampling brings $\bm{M}_t^K$ closer to $\bm{M}_t^0$.
However, estimating the observation-dependent distribution is computationally complex due to the high dimensionality and complex dependencies.

To avoid this complexity, our method simplifies the anisotropic Gaussian distribution as a distribution with a diagonal covariance.
This simplification is necessary because the covariance matrix grows quadratically with the prediction length $T_p$, making it computationally expensive for large $T_p$.
This simplified anisotropic Gaussian distribution is represented by two lightweight neural networks, $\bm{\mu}_{\eta}$ and $\bm{\sigma}_{\xi}$ where $\eta$ and $\xi$ are the parameters of the neural networks, conditioned on observation.
With $\bm{\mu}_{\eta}$ and $\bm{\sigma}_{\xi}$, the anisotropic Gaussian distribution is represented as $\mathcal{N}(\bm{\mu}_{\eta}(\bm{X}_t),\bm{\sigma}_{\xi}(\bm{X}_t))$,

\vskip\baselineskip
\noindent
\textbf{(b) Lightweight network for $\mathcal{N}(\bm{\mu}_{\eta}(\bm{X}_t),\bm{\sigma}_{\xi}(\bm{X}_t))$:}

To clarify the key advantages of our lightweight network, it is compared with the most similar one, called BRIDGER~\cite{bridger_chen2024behavioral}. The architectures of BRIDGER and our NO-Diffusion are shown in Fig.~\ref{fig: intro} (b) and Fig.~\ref{fig: intro} (c), respectively. In NO-diffusion, (1) the observation encoder $E_{\psi}$ is shared with NOAM and the diffusion model, and (2) simple two-layer MLPs take $\bm{X}_{t}$ to estimate $\bm{\mu}_{\eta}$ and $\bm{\sigma}_{\xi}$. On the other hand, BRIDGER has two independent deeper encoders, leading to high computational cost and processing time, especially for high-dimensional data such as images.
We experimentally found that sharing $E_{\psi}$ not only enhances the processing speed but also boosts the accuracy of estimating the sampling distribution; see Table~\ref{tab:ablation}.
While the two-layer MLP is limited in its ability to represent complex distributions, it is sufficient for estimating a distribution with a diagonal covariance.

\begin{figure*}[!ht]
\centering
  \begin{minipage}{0.325\linewidth}
    \centering
    \includegraphics[width=\linewidth]{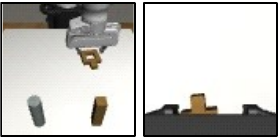}
    \subcaption{Square}
  \end{minipage}
  \begin{minipage}{0.325\linewidth}
    \centering
    \includegraphics[width=\linewidth]{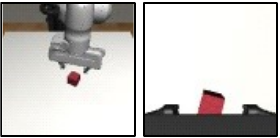}
    \subcaption{Lift}
  \end{minipage}
  \begin{minipage}{0.325\linewidth}
    \centering
    \includegraphics[width=\linewidth]{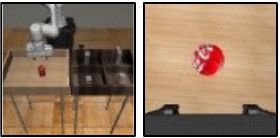}
    \subcaption{Can}
  \end{minipage}
  
  \caption[Tasks included in the robomimic dataset.]{Tasks included in the robomimic dataset~\cite{robomimic2021} along with third-person view images (left) and first-person view images (right) for each task. The model is trained using these two types of images along with the robot's position as input. Although robomimic contains five different tasks, this study focuses on three tasks: Square, Lift, and Can.}
  \label{fig:tasks}
\end{figure*}

\subsection{Training}
\label{subsection: training}

The observation encoder $E_{\psi}$, the mean estimator $\bm{\mu}_{\eta}$, the variance estimator $\bm{\sigma}_{\xi}$, and the diffusion model $\epsilon_{\theta}$ are trained jointly in an end-to-end manner.  
The total loss is as follows: 
\begin{equation}
    \mathcal{L}_{total} = \mathcal{L}_{diff} + w_{\sigma}\mathcal{L}_{\sigma} + w_{\mu}\mathcal{L}_{\mu},
    \label{equation: total}
\end{equation}
where $\mathcal{L}_{diff}$ denotes the diffusion loss, $\mathcal{L}_{\sigma}$ and $\mathcal{L}_{\mu}$ are auxiliary losses for training NOAM.
$w_{\sigma}$ and $w_{\mu}$ are weighting hyperparameters.
$\mathcal{L}_{diff}$ is defined as:  
\begin{equation}
    \mathcal{L}_{diff} = {\rm MSE}(\bm{M}_t^0, \epsilon_{\theta}(\bm{M}_{t}^k,\bm{X}_t, k)),
    \label{equation: l_diff}
\end{equation}
where $\bm{M}_{t}^k \sim \mathcal{N}(\bm{\mu}_{\eta}(\bm{X}_t), \bm{\sigma}_{\xi}(\bm{X}_t))$ and $\bm{X}_t = E_{\psi}(\bm{O}_t)$. $\bm{M}_{t}^k$ is sampled through the reparameterization trick.
This trick allows $\bm{\mu}_{\eta}$ and $\bm{\sigma}_{\xi}$ to be trainable from $\mathcal{L}_{diff}$, optimizing the distribution $\mathcal{N}(\bm{\mu}_{\eta}(\bm{X}_t), \bm{\sigma}_{\xi}(\bm{X}_t))$ to facilitate denoising by $\epsilon_{\theta}$. However, relying solely on $\mathcal{L}_{diff}$ makes training unstable. To address this, the auxiliary losses $\mathcal{L}_{\mu}$ and $\mathcal{L}_{\sigma}$ stabilize the training:  
\begin{eqnarray}
    \mathcal{L}_{\mu} &=& {\rm MSE}(\bm{M}_t^0, \mu_{\eta}(\bm{X}_t)),\\
    \mathcal{L}_{\sigma} &=& {\rm tr}((\bm{M}_t^{0} - \bm{\mu}_{\eta})(\bm{M}_t^{0} - \bm{\mu}_{\eta})^{\rm T} - \bm{\sigma}_{\xi}\bm{\sigma}_{\xi}^{\rm T}).
    \label{equation: sigma}
\end{eqnarray}
In Eq.~(\ref{equation: sigma}), $\bm{M}_t^{0}$, $\bm{\mu}_{\eta}$, and $\bm{\sigma}_{\xi}$ are reshaped from tensors to vectors for implementation purposes.
$\bm{X}_t$ is omitted for brevity in Eq.~(\ref{equation: sigma}). 
Estimation of the mean and variance of $\bm{M}_t^{0}$ prevents $\bm{\mu}_{\eta}$ and $\bm{\sigma}_{\xi}$ from getting stuck at local optimum. Since the true mean is unknown, $\bm{\mu}_{\eta}$ is used in Eq.~(\ref{equation: sigma}) to compute the variance of $\bm{M}_t^{0}$.

\section{Experimental Results}
\label{section: experiments}

In this section, we evaluate the effectiveness of our No-Diffusion for robot motion planning, verifying its ability to generate accurate motions with minimal diffusion steps. As shown in Fig.~\ref{fig:tasks}, we conducted a series of experiments in a simulation environment, enabling reproducible and fair comparison.

The goal is to verify whether or not one-step diffusion with noise optimization is comparable to state-of-the-art multi-step methods. Specifically, we aim to investigate the following key questions:
\begin{enumerate} 
    \item{Can our method achieve motion planning performance comparable to multi-step diffusion approaches while significantly reducing computational cost? We address this question through quantitative evaluations in Sections~\ref{exp:qant_eval}, \ref{exp:compe_time}, and \ref{exp:qality_limit}.}
    \item{How does noise optimization affect motion accuracy and robustness? This is verified in the ablation study presented in Section~\ref{exp:ablation}.}
\end{enumerate}

To answer these questions, our experiments quantitatively evaluate both the motion accuracy and computational efficiency under diverse task conditions, which are evaluated by the task success rate and the runtime, respectively.
we compare our method against state-of-the-art methods, including Diffusion Policy~\cite{Diffusion_Policy}, Crossway Diffusion~\cite{crossway_diffusion}, and BRIDGER~\cite{bridger_chen2024behavioral}, with multiple evaluation metrics.
All these three methods are provided by their publicly available official codes.

\subsection{Dataset}

Our experiments are conducted with the robomimic dataset~\cite{robomimic2021}, which is well suited for systematic evaluation. It provides large-scale expert demonstrations across multiple manipulation tasks in a simulated environment built on RoboSuite~\cite{robosuite2020}, enabling controlled comparison, high-throughput sampling, and robust assessment of generalization to diverse object configurations.

Three robotic manipulation tasks are used in our experiments: 
Square, Lift, and Can, using the ``ph'' subset, which contains 200 expert demonstrations per task, ensuring that all motions successfully achieve the tasks.
These three tasks are shown in Fig.~\ref{fig:tasks}. Square involves inserting a wooden peg into a matching slot on a cube. The Lift task requires the robot to grasp and lift a red cube from the table. In Can, the robot picks up a red can from the left side of the table and places it into a designated frame on the right.
To evaluate the ability to generalize across different initial object placements, their placements differ among demonstrations.

The number of images, $N_{i}$ in Sec.~\ref{section: notation}, is two.
One image is captured by a camera attached to the end effector.
This camera observes the detail around the end effector.
The other camera is fixed in front of a robot to observe the entire workspace.
In our experiments, the size of the image is $H = W = 84$ pixels.

\subsection{Training Details}

Following the training and experimental protocol of~\cite{Diffusion_Policy,crossway_diffusion,bridger_chen2024behavioral}, all diffusion models, including our NO-Diffusion, are implemented as follows.
Each diffusion model consists of 100 diffusion steps (denoted by $K_{m}$), while the number of diffusion steps used at inference (i.e., $K$) varies between 1 and 100.
The diffusion models are trained across the full range of the diffusion steps from 1 to 100. The noise variance follows the DDIM diffusion scheduler~\cite{ddim}.

All models are trained with 100,000 iterations.

\subsection{Quantitative Evaluation}\label{exp:qant_eval}

\begin{table*}[t]
    \centering
    \caption{Success rate comparison with varying $K$ on the three tasks. The best score is highlighted in \red{red}, while the second-best score is shown in \blue{blue}.}
    \label{tab:comparison}
    \begin{tabular}{l|cccc|cccc|cccc}
        \toprule
        \multicolumn{1}{c|}{Task} & \multicolumn{4}{c|}{Square} & \multicolumn{4}{c|}{Lift} & \multicolumn{4}{c}{Can} \\
         \cmidrule{1-1} \cmidrule(lr){2-5} \cmidrule(lr){6-9} \cmidrule(lr){10-13}
        % Inference diffusion step 
            \multicolumn{1}{c|}{$K$} & 100 & 50 & 10 & 1 & 100 & 50 & 10 & 1 & 100 & 50 & 10 & 1 \\
        \midrule
        Diffusion policy~\cite{Diffusion_Policy}
         & \red{93.3} & \red{90.9} & 0.0 & 0.0 & \red{100.0} & \red{100.0} & 4.3 & 3.6 & \red{99.6} & \red{99.8} & 0.0 & 0.0 \\
        Crossway Diffusion~\cite{crossway_diffusion} & \blue{91.4} & 89.9 & 0.0 & 0.0 & \red{100.0} & \red{100.0} & 5.3 & 4.1 & \blue{99.1} & \blue{99.3} & 0.0 & 0.0 \\
        BRIDGER~\cite{bridger_chen2024behavioral} & 88.4 & 90.5 & 89.1 & 87.6 & 99.9 & 99.9 & 99.8 & 99.7 & 96.9 & 95.8 & 97.0 & 96.0 \\
        \midrule
        NO-Diffusion & 91.2 & 90.1 & \blue{90.6} & \red{90.8} & \red{100.0} & \red{100.0} & \red{100.0} & \blue{99.9} & 98.4 & 98.6 & \red{98.1} & \red{98.1} \\
        NO-Diffusion$^{\dagger}$ & 91.2 & \blue{90.8} & \red{91.6} & \red{90.8} & \red{100.0} & 99.9 & \red{100.0} & \red{100.0} & 98.4 & 98.1 & \blue{97.6} & \blue{97.1} \\
        \bottomrule
    \end{tabular}
\end{table*}

Table~\ref{tab:comparison} presents the task success rates for the three tasks.
In all quantitative performance comparisons shown in this paper (i.e., in Tables~\ref{tab:comparison}, \ref{tab:generate_action_time}, and \ref{tab:ablation}), the task success rate is the score averaged over 1,000 trials per task, with initial object placements varying.
At inference, each method is evaluated with varying $K = \{ 100, 50, 10, 1 \}$.
Additionally, we introduce Ours$^{\dagger}$, in which $K_{m} = K$, to verify whether or not the efficient model (i.e., $K_{m} = K$) improves the task success rate, while diffusion steps between $K_{m}$ and $K+1$ are not used at inference in Ours.

\noindent\textbf{Diffusion Policy and Crossway Diffusion:}  
Both methods achieve high success rates when $K = 100$ and $K = 50$. However, their performance drops drastically at $K = 10$ and $K = 1$, where the success rate reaches 0. This reveals their inability to generate appropriate actions with limited inference steps.

\noindent\textbf{BRIDGER:}  
BRIDGER achieves higher success rates than Diffusion Policy and Crossway Diffusion when $K$ is small (i.e., $K \leq 10$). This advantage stems from the use of a pre-trained action generator that supports accurate action generation even with limited denoising.
However, BRIDGER with $K=1$ drops to 87.6 on Square. This is an essential difference from NO-Diffusion, which is introduced below.

\noindent\textbf{NO-Diffusion (our method):}  
NO-Diffusion demonstrates strong robustness under limited inference steps. 
Although NO-Diffusion is slightly inferior to Diffusion Policy at $K=100$ and $K=50$, NO-Diffusion cannot work at all at $K=100$ and $K=50$.
Furthermore, NO-Diffusion outperforms BRIDGER across most $K$ values, particularly in smaller $K$. 
Notably, at $K=1$, NO-Diffusion achieves 90.8 on Square, slightly improving over its $K=10$ result (i.e., 90.6).
This result proves the ability of our NO-Diffusion to achieve one-step diffusion.

\noindent\textbf{NO-Diffusion$^{\dagger}$:}   
While the primary goal of this work is one-step diffusion for real-time robot control, NO-Diffusion$^{\dagger}$ is slightly improved from NO-Diffusion at $K=50$ and $K=10$.
This result demonstrates that not only the memory efficiency but also the motion planning is also improved by reducing $K_{m}$ to $K$.

Overall, our NO-Diffusion consistently performs well across different tasks, particularly under low inference steps $K$, in particular $K=1$, demonstrating its efficiency and adaptability.

\begin{table}[t]
  \centering
    \caption{Comparison of the motion prediction time and the success rate for the Square task. In Condition I, Diffusion Policy and Crossway Diffusion are executed with $K$ used in their papers. Since BRIDGER is tested with varying $K$ in its paper, $K=10$ is selected as a small value for efficient diffusion. In Condition II, $K$ of all these methods are selected so that the success rate is similar to that of our NO-Diffusion (i.e., 90.8).}
    \begin{tabular}{c|l|ccc}
    \toprule
    Condition & Method & $K$ & Time [s] & Success rate \\
    \midrule
    \multirow{3}[2]{*}{I} & Diffusion Policy & 100 & 2.691 & 93.3 \\
      & Cross Diffusion & 100 & 2.237  & 91.4 \\
      & BRIDGER & 10 & 0.500  & 89.1 \\
    \midrule
    \multirow{4}[4]{*}{I\hspace{-1.2pt}I} & Diffusion Policy & 50 &  1.297 & 90.9 \\
      & Cross Diffusion & 50 & 1.150 & 91.4 \\
      & BRIDGER & 50 & 2.346  & 90.5 \\
\cmidrule{2-5}      & NO-Diffusion (Ours) & 1 & 0.052 & 90.8 \\
    \bottomrule
    \end{tabular}%
  \label{tab:generate_action_time}%
\end{table}%

\subsection{Comparison of Motion Prediction Time}\label{exp:compe_time}

To evaluate the practicability of our approach, the motion prediction time should be shorter. We measure the inference time from receiving the observation \( \bm{O}_t \) to producing the output motion using an NVIDIA TITAN RTX. The inference time of our method is just 0.052 seconds with $K=1$, achieving a high success rate of 90.8\% with minimal computation. The results are presented in Table~\ref{tab:generate_action_time}.
Success rates in Table~\ref{tab:generate_action_time} are reported as the average over the three manipulation tasks.

\begin{table}[t!]
    \centering
    \renewcommand{\arraystretch}{1.2}
    \setlength{\tabcolsep}{4pt}
    \caption{Ablation study on key design choices. Each setting is evaluated in terms of task success rate.}
    \label{tab:ablation}
    \begin{tabular}{l|cccc|ccc|c}
    \toprule
      & \multicolumn{4}{c|}{Condition} & \multicolumn{1}{c}{Success Rate} \\
      \cline{2-6}
    Setup & $\mathcal{L}_{total}$ & $E_{\psi}$ & $\bm{\sigma}_{\xi}$ & $\bm{\mu}_{\eta}, \bm{\sigma}_{\xi}$ & \multicolumn{1}{c}{Average} \\
    \midrule
    \rowcolor[rgb]{ .949,  .949,  .949} \textbf{I} & Eq. (\ref{equation: total}) & Shared & $\mathbb{R}^{10}$ & $\bm{\mu}_{\eta}(\bm{X}_t), \bm{\sigma}_{\xi}(\bm{X}_t)$ & 96.3 \\
    I\hspace{-1.2pt}I & $\mathcal{L}_{diff}$ & Shared & $\mathbb{R}^{10}$ & $\bm{\mu}_{\eta}(\bm{X}_t), \bm{\sigma}_{\xi}(\bm{X}_t)$ & 96.0 \\
    I\hspace{-1.2pt}I\hspace{-1.2pt}I & Eq. (\ref{equation: total}) & Separated & $\mathbb{R}^{10}$ & $\bm{\mu}_{\eta}(\bm{X}_t), \bm{\sigma}_{\xi}(\bm{X}_t)$ & 95.9 \\
    I\hspace{-1.2pt}V & Eq. (\ref{equation: total}) & Shared & $\mathbb{R}^{1}$ & $\bm{\mu}_{\eta}(\bm{X}_t), \bm{\sigma}_{\xi}(\bm{X}_t)$ & 95.9 \\
    V & Eq. (\ref{equation: total}) & Shared & $\mathbb{R}^{10}$ & Const. & 95.2 \\
    \bottomrule
    \end{tabular}
    \vspace{-1em}
\end{table}

Under Condition I, all state-of-the-art methods are evaluated with the $K$ originally adopted in their papers, except for BRIDGER. For BRIDGER, although the original paper reports results with multiple values of $K$ such as 0, 5, 20, 80, and 160, we adopt the result with $K=10$, which is closest to the smallest reported values (5 and 20) and aligns with the paper's emphasis on achieving high performance with fewer diffusion steps.
These settings yield relatively high success rates with the expense of efficiency. For example, Diffusion Policy requires 2.69 seconds to achieve 93.3\% success, and Cross Diffusion needs 2.24 seconds for 91.4\%. BRIDGER achieves 89.1\% success in 0.50 seconds under $K=10$, showcasing low latency but slightly lower performance. In contrast, our method achieves a comparable success rate of 90.8\% using only a single inference step, resulting in an inference time that is approximately 51.8$\times$ faster than Diffusion Policy, 43.0$\times$ faster than Cross Diffusion, and 9.6$\times$ faster than BRIDGER.

Under Condition II, $K$ of each state-of-the-art method is tuned to a configuration where its success rate is closest to that of NO-Diffusion. Even in this setting, our method outperforms all baselines in terms of efficiency. For example, Diffusion Policy and Cross Diffusion require 1.30 and 1.15 seconds, respectively, to reach around 91\% success. BRIDGER needs to increase $K$ to 50 to achieve a comparable success rate of 90.5\%, which results in a computation time of 2.35 seconds. In comparison, the inference time of NO-Diffusion is 0.052 seconds, achieving speed-ups of 24.9$\times$, 22.1$\times$, and 45.1$\times$, respectively.

These results confirm that our one-step diffusion policy achieves near-optimal task performance at a fraction of the computational cost, making it especially well-suited for real-time robotic applications.

\subsection{Ablation Study}\label{exp:ablation}

As shown in Table~\ref{tab:ablation}, ablation experiments are conducted to evaluate the contribution of each key component in our method.
Setup~I is our full method.
Success rates are reported as the average over the three manipulation tasks.

\noindent\textbf{Auxiliary Losses (Setup~II):} Ablating the auxiliary losses \( \mathcal{L}_{\mu} \) and \( \mathcal{L}_{\sigma} \) slightly reduces the overall performance (i.e., from (I) 96.3 to (II) 96.0). These terms are introduced to stabilize training by regularizing the estimated mean and variance. While the decrease in the average success rate is modest (i.e., 0.3 points), the results indicate that these losses help prevent overfitting and improve generalization, particularly in the Square task.

\noindent\textbf{Separated Encoder (Setup~III):} The encoders for the diffusion model and the noise estimator are separated, following the design used in BRIDGER.
We can see that the success rate improves by sharing the encoders in our method (i.e., from (III) 95.9 to (I) 96.3).
Furthermore, our shared encoder design improves computational efficiency.

\noindent\textbf{Isotropic Gaussian (Setup~IV):} This setting uses an observation-dependent isotropic Gaussian instead of an anisotropic Gaussian.
The success rate improves by sharing the encoders in our method (i.e., from (III) 95.9 to (I) 96.3).
We interpret that the anisotropic distribution can improve the performance because modeling per-dimension variance is important for capturing the diverse nature of robot actions, which include position, orientation, and gripper control.

\noindent\textbf{Static Sampling Distribution (Setup~V):} Unlike our original method, $\bm{\mu}_{\eta}$ and $\bm{\sigma}_{\xi}$ in NOAM are trained without taking $\bm{X}_{t}$.
That is, NOAM estimates the sampling distribution independently of the observation $\bm{O}_{t}$.
Specifically, $\bm{\mu}_{\eta}$ and $\bm{\sigma}_{\xi}$ starts with trainable $\bm{\mu}$ and $\bm{\sigma}$, respectively.
These $\bm{\mu}$ and $\bm{\sigma}$ are trained with $\bm{\mu}_{\eta}$ and $\bm{\sigma}_{\xi}$ by backpropagation.
This simplification reduces the success rate (i.e., from (I) 96.3 to (V) 95.2). This result demonstrates the effectiveness of dynamic observation-dependent sampling noise estimation.

\begin{figure*}[t]
    \centering
    \begin{subfigure}[b]{0.32\linewidth}
        \centering
        \includegraphics[width=\linewidth]{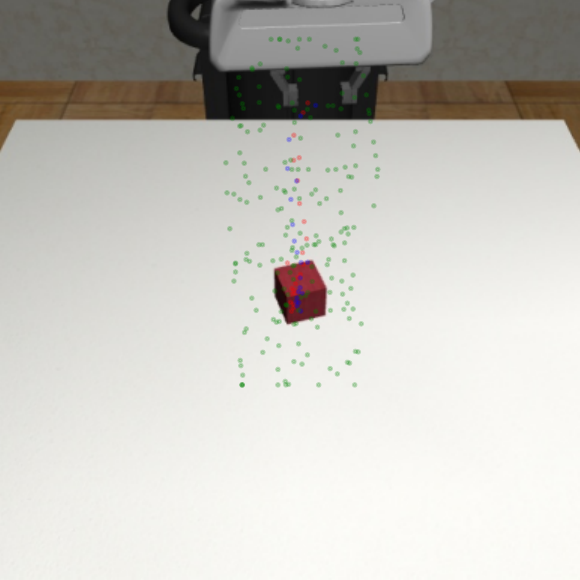}
        \caption{Lift: \red{Success}, \blue{Success}, \pltgreen{Failure}}
    \end{subfigure}
    \hfill
    \begin{subfigure}[b]{0.32\linewidth}
        \centering
        \includegraphics[width=\linewidth]{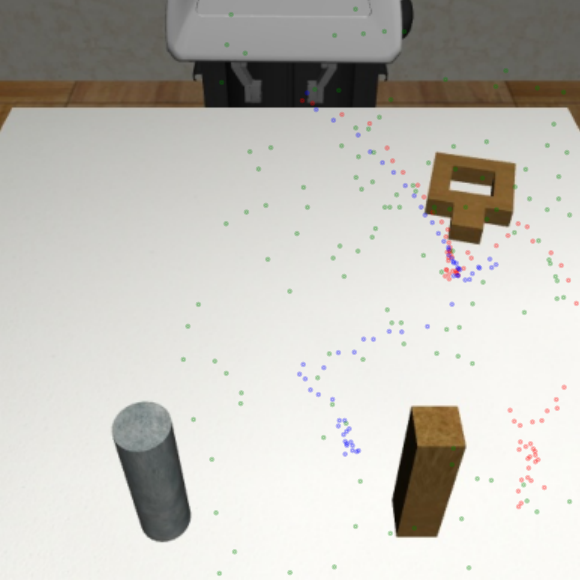}
        \caption{Square: \red{Success}, \blue{Success}, \pltgreen{Failure}}
    \end{subfigure}
    \hfill
    \begin{subfigure}[b]{0.32\linewidth}
        \centering
        \includegraphics[width=\linewidth]{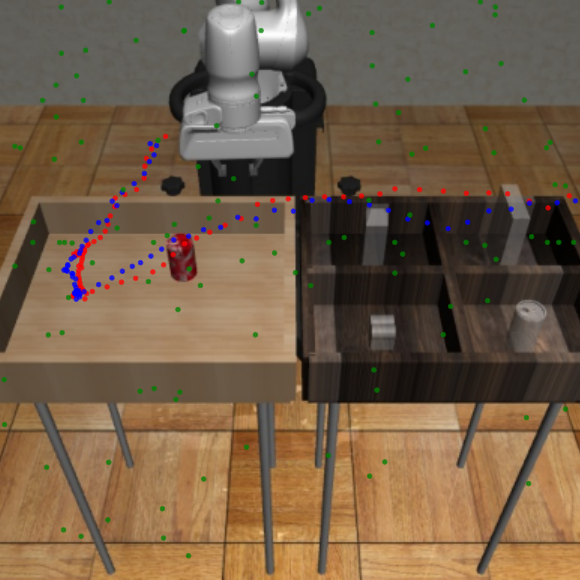}
        \caption{Can: \red{Success}, \blue{Success}, \pltgreen{Failure}}
    \end{subfigure}\\
    \begin{subfigure}[b]{0.32\linewidth}
        \centering
        \includegraphics[width=\linewidth]{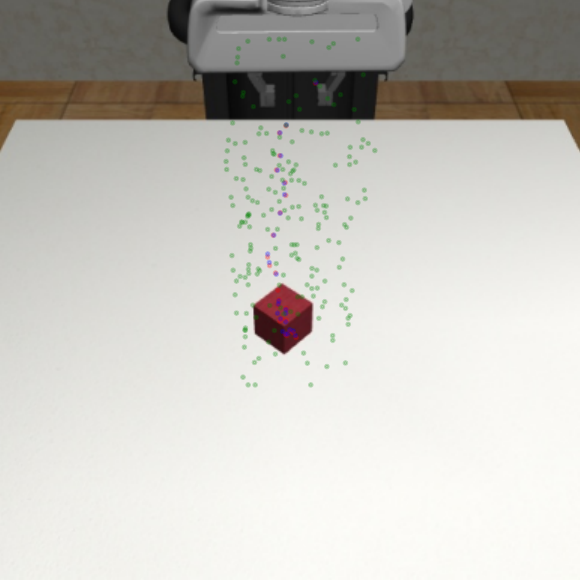}
        \caption{Lift: \red{Success}, \blue{Success}, \pltgreen{Failure}}
    \end{subfigure}
    \hfill
    \begin{subfigure}[b]{0.32\linewidth}
        \centering
        \includegraphics[width=\linewidth]{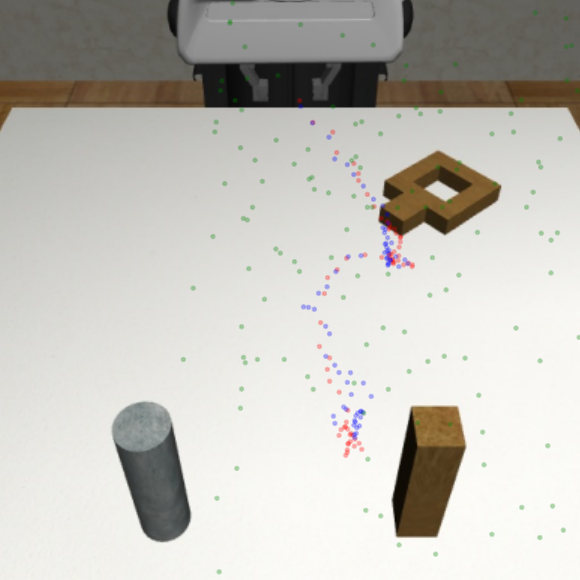}
        \caption{Square: \red{Success}, \blue{Success}, \pltgreen{Failure}}
    \end{subfigure}
    \hfill
    \begin{subfigure}[b]{0.32\linewidth}
        \centering
        \includegraphics[width=\linewidth]{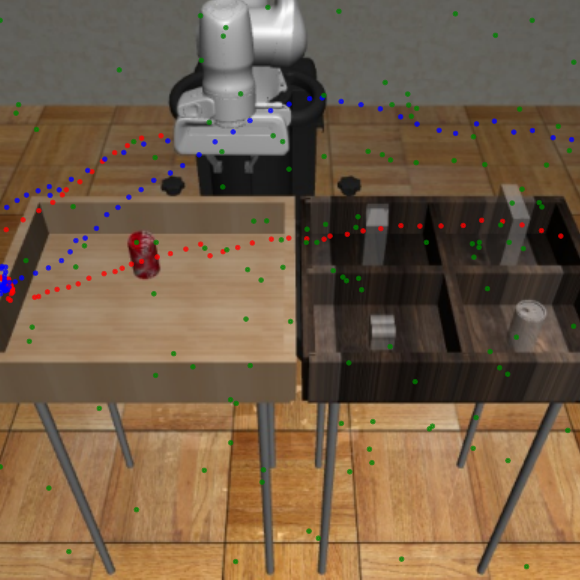}
        \caption{Can: \red{Failure}, \blue{Failure}, \pltgreen{Failure}}
    \end{subfigure}
    \vspace{-0.2em}
    \caption{
        Qualitative results of the three methods.
        Red, blue, and green dots indicate the motion trajectories of our \red{NO-Diffusion ($K=1$)}, \blue{Diffusion Policy ($K=100$)}, and \pltgreen{Diffusion Policy ($K=1$)}, respectively.
        The caption of each example tells whether the task is successful or failed.
        The color of the caption indicates the method.
    }
    \label{fig: qualitative_3examples}
    \vspace{-1.5em}
\end{figure*}

\subsection{Qualitative Evaluation and Limitation}\label{exp:qality_limit}

Examples of predicted motions are visualized in Fig.~\ref{fig: qualitative_3examples}.
In all examples, Diffusion Policy with 1 step produces almost random motions because one-step diffusion is insufficient for Diffusion Policy.
In the Lift and Square tasks, our NO-Diffusion and Diffusion Policy with 100 steps succeeded in the task, while Diffusion Policy with 1 step failed.
The motions of NO-Diffusion and Diffusion Policy with 100 steps are similar between the upper and lower examples probably because motions for grasping a cube have less freedom.
On the other hand, the motions of NO-Diffusion and Diffusion Policy with 100 steps are totally different in the upper example of the Square task.
This example reveals the necessity of probabilistic motion prediction, in which different motions can be predicted stochastically.
In the Can task, a robot picks the can up, moves it to the basket, and puts it in the basket.
Since the Can task is more difficult than the other two tasks, all three methods failed the task in the lower example.
However, even in this failure example, while the motion of Diffusion Policy swerves from the path toward the basket, the motion of NO-Diffusion moves toward the basket successfully. 
A supplementary video showing these qualitative results is available\footnote{\url{https://drive.google.com/file/d/1Bop_5EUN0GVVCTtNnyf3Y3oG8devlnvE/view?usp=sharing}}.

%%%%%%%%%%%%%%%%%%%%%%%%%%%%%%%%%%%%%%%%%%%%%%%%%%%%%%%%%%%%%%%%%%%%%%

\section{Conclusion}
\label{section: conclusion}

This paper proposed an image-based robot motion planning method using one-step diffusion, which can reduce the motion prediction time to 0.052 seconds.
Compared with state-of-the-art methods, our method can maintain high success rates even with one-step diffusion.

Future work includes more comprehensive experiments to verify the robustness of our method against diverse conditions, such as a variety of different tasks, different camera positions, and different robots.
The sim-to-real issue~\cite{robust_visual_sim2real,simtoreal_via_simtosim,i-sim2real} is also essential to achieve applications in the real world.

\bibliographystyle{ieee_fullname}
\bibliography{main}

\end{document}